\documentclass[letterpaper, 10 pt, conference]{ieeeconf} 
\IEEEoverridecommandlockouts 

\usepackage[ruled,vlined,linesnumbered]{algorithm2e}

\usepackage[cmex10]{amsmath}
\usepackage{amssymb,mathtools}
\usepackage{color, soul}
\usepackage{graphicx}
\usepackage{wrapfig}
\usepackage{verbatim} %,url}
\usepackage{enumerate}
\usepackage{pifont}
\usepackage{hyperref}
\usepackage{svg}

\long\def\omit#1{}

\newcommand{\sq}{\hbox{\rlap{$\sqcap$}$\sqcup$}}
\newcommand{\qed}{\hspace*{\fill}\sq}

\usepackage[linesnumbered, ruled,vlined]{algorithm2e}

\usepackage{tabularx}
\newcounter{protocol}

\newcommand{\eqsize}{\small}

\begin{document}
\title{\LARGE \bf Towards Optimal Human-Robot Interface Design Applied to Underwater Robotics Teleoperation
}

\author{Paulo Padrao, Jose Fuentes, Tero Kaarlela, Alfredo Bayuelo, Leonardo Bobadilla
\thanks{L. Bobadilla, P. Padrao, and J. Fuentes are with the School of Computing and Information Sciences, Florida International University, Miami, FL 33199, USA.
{\tt\small bobadilla@cs.fiu.edu, ppadraol@fiu.edu, jfuen099@fiu.edu}
A. Bayuelo is with the National University of Colombia, Bogotá,
Colombia. {\tt\small ajbayuelos@unal.edu.co} 
T. Kaarlela is with Centria University of Applied Sciences, 84100 Ylivieska, Finland.
{\tt\small tero.kaarlela@centria.fi}}}

\maketitle
\begin{abstract}
Efficient and intuitive Human-Robot interfaces are crucial for expanding the user base of operators and enabling new applications in critical areas such as precision agriculture, automated construction, rehabilitation, and environmental monitoring. In this paper, we investigate the design of human-robot interfaces for the teleoperation of dynamical systems. The proposed framework seeks to find an optimal interface that complies with key concepts such as user comfort, efficiency, continuity, and consistency. As a proof-of-concept, we introduce an innovative approach to teleoperating underwater vehicles, allowing the translation between human body movements into vehicle control commands. This method eliminates the need for divers to work in harsh underwater environments while taking into account comfort and communication constraints. We conducted a study with human subjects using a head-mounted display attached to a smartphone to control a simulated ROV. Also, numerical experiments have demonstrated that the optimal translation is often the most intuitive and natural one, aligning with users' expectations.
\end{abstract}

\section{Introduction}
\label{sec:introduction}
The design of intuitive, robust, and efficient Human-Robot interfaces (HRI) can extend the user base of operators and enable a new generation of applications in critical areas such as precision agriculture, automated construction, rehabilitation, and environmental monitoring \cite{hauser2013}.
Due to advances in virtual reality applications in the last few years, user interface design for robotic teleoperation drew increasingly more attention since the human performance of teleoperated systems can be decremented by data bandwidth, time delays, frame rates, and lack of concentration, among other interface-related factors\cite{chen2007}. Furthermore, environments with high spatiotemporal variability, sensing, and communication challenges, such as marine environments, provide additional challenges for teleoperation. 

Approximately $60\%$ of the global population resides within a distance of fewer than 100 kilometers from a coastal or estuary environment \cite{cohen97}. As these regions' population density and economic activity grow, the coastal ecosystems' stressors also increase. Therefore, underwater robots are increasingly essential for studying, monitoring, and managing coastal and estuarine environments. These environments are critical for numerous applications, including coastal conservation \cite{terraciano2020}, coral restoration \cite{quattrini2016data}, prediction of marine phenomena of interest \cite{padrao2022}, and oil rig maintenance \cite{shukla2015}.

As an example in coastal conservation and restoration, underwater robots can assist in monitoring marine protected areas to ensure compliance with regulations and detect illegal activities. In coral repair, robots can aid in planting new corals and removing harmful debris from damaged reefs. Meanwhile, in oil rigs, robots can inspect and maintain pipelines, ensuring the safety of workers and reducing the risk of oil spills. With their ability to collect high-quality data and perform tasks in hazardous environments, underwater robots have the potential to significantly improve our knowledge of these critical environments and support a range of applications.

The underwater environment is unnatural for humans, and working below the surface requires diving gear or teleoperated underwater robots. This paper presents a novel concept for teleoperating a remotely operated underwater vehicle (ROV). The proposed concept breaks the barrier between the underwater location and the user, freeing the user from harsh underwater conditions. The approach enables the remote operator to control an ROV with his body and head movements. Furthermore, a head-mounted display (HMD) enables the teleoperation of the ROV and provides visual feedback of the underwater world.

The teleoperation method presented is an intuitive and natural high-level control interface of the ROV for the human operator. The connection link between the ROV and the teleoperator utilizes an Ethernet connection over the Internet, enabling location-independent teleoperation of the ROV. Furthermore, the presented solution uses a smartphone to translate teleoperator head movements into ROV control commands, allowing a high-level control interface. The industry can benefit from the solution, particularly in conducting underwater tasks such as inspection, welding, and rescue. By operating remotely, the teleoperator can ensure an ergonomic and safe working environment, thereby reducing the risk of physical strain or injury. Additionally, this solution can allow for continuous work without frequent breaks or short working periods, ultimately increasing productivity and efficiency in underwater operations.

The primary contributions of this paper are as follows:

\begin{itemize}
\item We formulate an optimization-based framework for human-robot interface design that complies with user comfort and efficiency constraints. 
\item We present a novel concept for the teleoperation of an ROV by capturing and translating human body motions to control commands of the ROV.
\item We present a prototype of the system as a proof of concept.
\end{itemize}

The rest of the paper is organized as follows: Section \ref{sec:related} reviews previous research on the topic. Section \ref{sec:problem_formulation} formulates the two research problems in this paper. Section \ref{sec:methods} presents the methods used to solve the proposed problems, and section \ref{sec:implementation} presents the implementation of the prototype teleoperation as a proof-of-concept. Section \ref{sec:results} presents the results of this work, and section \ref{sec:conclusions} concludes this paper.    

\section{Related Work} \label{sec:related}

Our work addresses the problem of {\em optimal human-robot interface design}. The design of human-robot interfaces relies on the selection of appropriate desired properties, such as consistency, linearity, and continuity, that enable comfort and naturalness in teleoperation \cite{mimnaugh2021}. Our approach is closely related to the concepts presented in \cite{hauser2014} in the sense of investigating the mappings between human and robot spaces and the mathematical formulation of such a problem. We also share commonalities with recent research on HMD-based immersive teleoperation interfaces  \cite{halkola2022}, human perception-optimized planning \cite{lavalle2019}, and approaches that utilize optimal control for teleoperating robots \cite{havoutis2017}.

Teleoperation has been researched for decades to overcome barriers between the teleoperator and the environment~\cite{HanbookOfRobotics, lee2006, kofman2005}. The barrier can be a physical barrier, such as a wall, or an environmental barrier, such as a toxic or hazardous environment or deep underwater conditions.

In recent years, research on underwater teleoperation has focused on developing techniques to improve the performance and capabilities of underwater robots \cite{moni2022}. Current examples of underwater teleoperation applications range from developing virtual reality interfaces for underwater missions \cite{cruz2020} and the design of underwater humanoid robots \cite{wu2020} to visible light communication systems that can be employed for limited-range teleoperation of underwater vehicles \cite{codd2018}.

Underwater teleoperation poses several challenges due to the harsh and variable nature of the environment. One of the primary challenges is the limited communication bandwidth and signal quality, which can result in time delays and packet losses. The physical properties of the water prevent utilizing radio waves as a wireless communication method of teleoperation. Instead, acoustic and optical communication has been researched and utilized~\cite{Sheridan1981}. Both enable wireless underwater communication with either low bandwidth or limited usability. Acoustic modems are limited in bandwidth, enabling only low-resolution and low-quality video transmission~\cite{Barbie2021}. While optical modems enable higher bandwidth than acoustic modems, the tradeoff is the requirement for line-of-sight between the transmitter and the receiver. Maintaining line-of-sight between the ROV and the above-surface transmitter requires sophisticated tracking electronics and actuators~\cite{Kaushal2016}. The lack of visual cues in the underwater environment makes it difficult for operators to perceive the robot's location and orientation accurately. Additionally, the effects of water currents and turbulence can make it challenging to control the robot's movements accurately.

\section{Problem Formulation}
\label{sec:problem_formulation}
In this paper, we consider the task of visual-based teleoperation of an underwater vehicle. The two agents involved, the person teleoperating the robot and the robot itself, have a workspace, an action space, and a state space. Let $\mathcal{W}_o \subset \mathbb{R}^3$ and $\mathcal{W}_r \subset \mathbb{R}^3$ be the workspaces for the operator and the robot, respectively; $\mathcal{C}_o$ and $ \mathcal{C}_r$ be their configuration spaces. We denote by $\mathcal{U}$ the set of controls applicable to the robot and $\mathcal{A}$ the set of actions the human operator can perform. Following the notation from \cite{hauser2014}, we assume that the robot's dynamics is ruled by the relation given by the function $f: \mathcal{C}_r \times \mathcal{U} \longrightarrow \mathcal{C}_r$

\begin{equation}
\eqsize
    \dot{x} = f(x,u).
\end{equation}

To establish a teleoperating system, it is necessary to map the user's actions to the robot's actions through a map $g: \mathcal{C}_r  \times \mathcal{A}  \longrightarrow \mathcal{U}$ so that the robot is affected by actions taken by the user

\begin{equation}
\eqsize
\label{control}
    \dot{x} = f(x,g(x,a)).
\end{equation}

Assuming fixed sets $\mathcal{A}$, $\mathcal{U}$ and the function $f$, our problem is building $g$ according to principles such as \textit{Continuity}, \textit{Consistency} and \textit{Reachability} as described in \cite{hauser2014} and restated below. However, we may face the challenge wherein the user becomes fully immersed in the teleoperating system, leading to a lack of comprehensive or precise knowledge about the configuration state of the robot. Additionally, the information provided by the teleoperating system may present challenges with regard to fast and precise user interpretation for optimal performance. Nevertheless, the function $g$ assumes that there is perfect information from the robot's configuration state. For this reason, we require a map $\psi : \mathcal{C}_o \longrightarrow \mathcal{C}_r$; this map will translate robot configuration states to user configuration states and vice-versa. In fact, we assume that $\psi$ is a diffeomorphism. In that case, we define $z$ by $\psi(z) = x $ and \eqref{control} changes to

\begin{equation}
\eqsize
\label{control 2}
\dot{z} = [D{\psi}(z)]^{-1}\cdot f(\psi(z),g(\psi(z),a)).
\end{equation}

This formulation allows considering the problem from the point of view of the person operating the robot. In this scenario, we assume that the robot can put in considerably more effort than the operator. This consideration allows the definition of functionals tailored to directly address issues related to the person, which may not be preserved by the function $\psi$. 
As mentioned before, to achieve a realistic and comfortable teleoperation experience, the map $g$ should fulfill key conditions \cite{hauser2014}, of which we consider four to be the most crucial. Firstly, \textit{Consistency}, which means preserving the attributes that the robot and the operator share. In particular, this implies symmetry, meaning that if the operator's and robot's actions are symmetric with respect to a particular axis, the map $f$ should preserve this symmetry as much as possible. Secondly, \textit{Continuity}, which requires mapping actions of the operator to closely related actions of the robot. In this case, the derivative of $f$ may also need to be restricted in order to prevent the robot from moving inconsistently due to sensory-motor aspects. Thirdly, \textit{Linearity}, which provides an intuitive way for the agent to teleoperate the robot. In this case, it is reasonable to expect that if the input is doubled, the operator expects the robot output to be approximately doubled as well. Lastly, \textit{Reachability} and \textit{Completeness} provide the user the ability to operate the robot to a given desired state. Let $\mathcal{F}_o \subseteq \mathcal{C}_o$ and $\mathcal{F}_r \subseteq \mathcal{C}_r$ be a subset of feasible configurations for the operator and robot, respectively. A state $x' \in \mathcal{F}_r$ is said to be $u$-reachable from $x \in \mathcal{F}_r$ if there exists a control function $u(t)$ such that $x$ is brought to $x'$. In the same way, a state $x' \in \mathcal{F}_o$ is said to be $a$-reachable from $x \in \mathcal{F}_o$ if there exists a control function $a(t)$ such that $x$ is brought to $x'$. Therefore, the function $g$ is said to be \textit{complete} if all $u$-reachable pairs $(x, x') \in \mathcal{F}_r$ have an equivalent $a$-reachable pair $(x, x') \in \mathcal{F}_o$.

In general, the problem is stated as finding a function $g$ such that it minimizes certain functional $J(g)$ tailored to address a specific problem. As a general rule, this function has the form

\begin{equation}
\eqsize
    J(g) = \int_0^T L(x, g, Dg)dt,
\end{equation}

where $L$ is a cost function and $D^n$ indicates the $n^{th}$ derivative of $g$. Sometimes, when required, $a$ can be found together $g$ so the optimization problem could have two variables, so we will assume that both will be required unless stated otherwise. Naturally, an optimal control problem appears with $g \in C^1(\mathcal{C}_r \times \mathcal{A}, \mathcal{U})$ and $a \in C^1(\mathcal{A})$  being the solution to the constrained problem 

\begin{equation}
\eqsize
\label{min fun}
\begin{aligned}
    \min_{g,a} \  &\displaystyle \int_0^T L(x, g, Dg)dt \\
    \text{s.t.} \ \ &\dot{x} =f(x,g(x,a))\\
    & x(0) =x_{initial}, \ x(T)=x_{final}.
\end{aligned}
\end{equation}

We emphasize that $x=x(t)$ and $a=a(t)$ have temporal dependencies, and we dropped them in \eqref{min fun} to ease the notation. Similarly, a functional $\hat{J}$ can be defined taking into account \eqref{control 2} to make explicit the dependency on the trajectory performed by the operator.

In any case, the solution to this problem results in solving a Jacobi-Bellman equation \cite{hauser2014}, which involves solving a partial differential equation to obtain optimal functions. Moreover, this procedure is, in general, challenging, and some solutions could not fulfill the principles described before or could be difficult for an average person to execute. Therefore, we will study particular cases that have an intuitive design and are easy to manage for an average person.

\noindent
\textbf{Problem 1: Translating configuration spaces}

\textit{Given the configuration space of the operator $\mathcal{C}_o$ and the set of actions $\mathcal{A}$ the operator can perform, compute the mapping $g$ that translates operator configuration and action spaces into robot action and configuration spaces $\mathcal{C}_r$ and $\mathcal{U}$, respectively.}

Underwater communication is subject to various constraints that can impact the reliability and speed of data transfer, such as attenuation, noise, interference, latency, and power consumption. Also, the available bandwidth for underwater communication is limited, which can make it difficult to transmit large amounts of data quickly \cite{zhu2020}. 
Because humans have an inherent ability to reconstruct geometry and meaning from low-quality images, such as blurry or pixelated images, and the underwater environment poses severe data communication constraints, we investigate the problem of translating observation spaces to enable efficient data exchange for human-robot interaction. Let $\mathcal{Y}_{r}$ and $\mathcal{Y}_{o}$ be the observation spaces of the robot and operator, respectively. We want to compute maps such that the transmitted information is small but meaningful enough to allow the operator to effectively and efficiently operate the robot. This is particularly useful in real-time scenarios in which high computational tasks are out of our scope. 

In this way, we define a functional to be optimized in order to find a suitable mapping. Let $h: \mathcal{Y}_{r}\longrightarrow \mathcal{Y}_{o}$ be such a mapping. Therefore, \textbf{Problem 1} can be formulated using another cost function $K$, and a function $h\in \mathcal{H}$ is to be found in a suitable function space, $\mathcal{H}=C^n(\mathcal{C}_r \times \mathcal{A}, \mathcal{U})$ for instance, such that

\begin{equation}
\eqsize
\label{first map h opt}
    \max_{h} K(h, Dh, D^2h, \ldots, D^nh).
\end{equation}

As discussed before, specific properties need to be fulfilled to make the information useful and not generate additional distortions. Thus, we borrow attributes from the ideal function $g$ presented before and \cite{hauser2014}, such as Continuity and Linearity; however, the linearity assumption can be relaxed to preserve affine transformations as the environment geometry is somehow preserved. Because we also face problems related to communication, we may consider $h$ as a time-dependent function  $h: \mathcal{Y}_{r}\times[0, T]\longrightarrow \mathcal{Y}_{o}$ that has a time constraint related to the communication threshold $c(t)$. Hence, \eqref{first map h opt} can be updated to 

\begin{equation}
\eqsize
    \begin{aligned}
\label{second map h opt}
    \displaystyle \max_{h}  \quad & K(h, Dh, D^2h, \ldots, D^nh)\\
    \text{s.t } \quad & I(h(y,t))\leq c(t) \ \text{for each} \ y\in \mathcal{Y}_r.
    \end{aligned}
\end{equation}

Where $I:\mathcal{Y}_o \longrightarrow \mathbb{R}$ computes the information sent to the operator at given instant $t$.

\noindent
\textbf{Problem 2: Translating observation spaces}

\textit{Given the observation space of the robot $\mathcal{Y}_r$, compute a mapping $h: \mathcal{Y}_r \longrightarrow \mathcal{Y}_o$ that gives enough information to the operator to effectively and remotely control the motion of the underwater robot.}

\section{Methods} 
\label{sec:methods}

\subsection{Problem 1: Translating Configuration Spaces}

To address the first problem of translating operator spaces into robot spaces, we examine the scenario wherein the human operator transmits commands to the robot using head movements and body motion. The range of motion for the human head depends on several factors, such as age, sex, health, and individual anatomical differences. On average, an adult human can rotate their head up to $90$ degrees to either side and tilt their head up and down about $45$ degrees, giving a total range of motion of about $180$ degrees. The movements mentioned above are captured by a smartphone's built-in sensors, including the inertial measurement unit (IMU) and barometer, which are affixed to a diving mask worn by the operator. The operator's movements are subsequently translated into commands to enable the teleoperation of the underwater robot. The complete architecture is shown in Fig. \ref{fig:architecture}. The visual feedback is provided to the operator through the robot camera, which captures images of the robot's environment and displays them on the smartphone screen. This allows the operator to see where the robot is going and adjust its movements accordingly. Due to the restrictions of user input commands, the robot is initially treated as a rigid body that moves at a constant speed in $\mathbb{R}^2$.
Let $a_\theta$ and $a_\psi$ be the head pitch and head yaw commands, respectively, and $u_\theta$ and $u_\psi$ be the robot camera tilt command and yaw command of the robot base. Let $v_o$ and $v_r$ be the linear forward velocities of the operator and the robot in $[m/s]$, respectively. Also, we consider that the depth of the operator and ROV can be set directly by the action variables $a_z$ and $u_z$ in $[m]$, respectively. We define the action space of the operator as 

\begin{multline}\label{operator_action_space_eq}
\eqsize
    \mathcal{A} =  (a_{\theta\min}, a_{\theta\max}) \times (a_{\psi\min}, a_{\psi\max}) \times (v_{o\min}, v_{o\max}) \\ \times (a_{z\min}, a_{z\max})
\end{multline}

and the action space of the robot as 
\begin{multline}\label{robot_action_space_eq}
\eqsize
     \mathcal{U} =  (u_{\theta\min}, u_{\theta\max}) \times (u_{\psi\min}, u_{\psi\max}) \times (v_{r\min}, v_{r\max}) \\ \times (u_{z\min}, u_{z\max})
\end{multline}

\begin{figure}[ht!]
    \centering
    \includegraphics[scale=1]{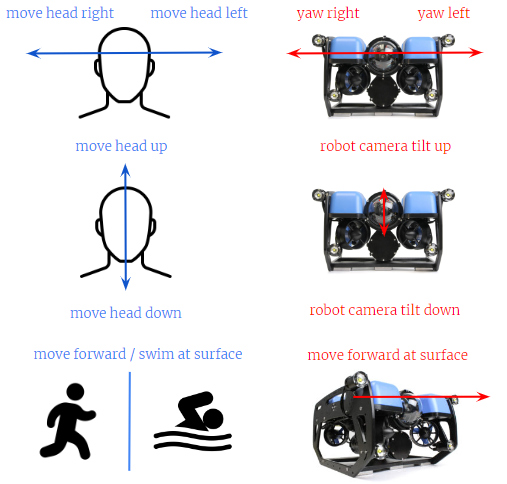}
    \caption{An example of human-robot action space translation}
    \label{fig:human_robot_config}
\end{figure}

An example of human-robot action space translation is shown in Fig. \ref{fig:human_robot_config}.

To evaluate the mappings, we modified functionals defined in \cite{hauser2014} to reflect specific factors that are relevant to underwater teleoperation. Proposed problems can be cast using formulation \eqref{control} or \eqref{control 2} when $\psi$ is available. The first formulation is the shortest-distance problem where 

\begin{equation}
\eqsize
\label{arc length}
    \begin{aligned}
     \min_{g,a} \ & \displaystyle \int_0^T L(x,g,Dg)dt =\displaystyle \int_0^T ||\dot{x}(t)|| dt \\
      = &\displaystyle \int_0^T ||f(x(t),g(x(t),a(t)))|| dt\\
     \text{s.t} \ \ & \dot{x}(t) = f(x(t),g(x(t),a(t)))\\
      & x(0) = x_{initial}, \ x(T)=x_{final}.
    \end{aligned}
\end{equation}

Several works have reported the need of maintaining a comfortable environment for the user during teleoperation tasks ~\cite{Becerra2020, lavalle2019}. In this sense, we seek to reduce the number of movements made by the person to increase the level of comfort when operating the robot. As an example, \cite{lavalle2019} relies on minimizing the number of head movements to improve the user's comfort. This can be seen as the simplification of the curvature of a curve. Intuitively, paths that do not have too many turns are considered. This is expressed by $\kappa(t) = ||\ddot{x}(t)||$, and the average curvature is given by

\begin{equation}
\eqsize
\label{curvature}
    \begin{aligned}
     \min_{g,a} \ & \displaystyle \int_0^TL(x,g,Dg) dx =\frac{1}{T}\displaystyle \int_0^T ||\ddot{x}(t)|| dt\\
     = & \frac{1}{T} \displaystyle \int_0^T ||f(f_x+f_ug_x) +f_u g_a\dot{a}|| dt\\
     \text{s.t.} \ \ & \dot{x}(t) = f(x(t),g(x(t),a(t)))\\
      & x(0) = x_{initial}, \ x(T)=x_{final}.
    \end{aligned}
\end{equation}

The function arguments were omitted to enhance readability, and the subscripts represent partial derivatives. It should be noted, however, that these functionals can be computationally expensive to evaluate. Furthermore, the norm function lacks a derivative at the origin. Despite this limitation, we still aim to utilize these functions as evaluators for translating actions between the human-robot action spaces.

Fig. \ref{fig:human_robot_config} depicts a mapping between the actions of the operators and the actions of the robot. This map exhibits key properties such as continuity and consistency, meaning that small operator movements result in corresponding small robot movements and that these actions are reversible and consistent. As a result, the robot is highly intuitive to control even though the robot's reachability is somewhat limited, as not all operator actions can be translated into robot actions. 

\subsection{Problem 2: Translating Observation Spaces}

To address the second problem of translating observation spaces, we consider that the observation space of the robot $\mathcal{Y}_r$ is composed only of video frames streamed from an RGB camera and can be defined as follows
\begin{equation}
    \mathcal{Y}_r = \mathbb{R}^{M \times N \times 3}\times [0,T]
    % \mathcal{Y}_{r}=[0,255]\times[0,255]\times[0,255]\times [0,M] \times [0,h] \times T
\end{equation}
where $M$ and $N$ are the width and height of the video frames, respectively, and $T$ is the timestamp. Because of communication constraints, the video frames from the robot are down-sampled and scaled to reduce their size while maintaining enough information to be processed by the operator. In terms of data transmission efficiency, black-and-white (binary) images are generally more efficient than grayscale or semantic segmentation since they only require one bit per pixel to represent the image.
% grayscale images require multiple bits per pixel, and edge detection or semantic segmentation require even more bits per pixel to represent the edges or segmentation masks.
The main tradeoff lies in the level of detail that can be represented in the image versus the amount of data required to represent it. A color image typically uses 24 bits per pixel to represent 256 levels of intensity at each channel. A gray-scale image considers just one channel, so it requires 8 bits per pixel, while a black-and-white image only uses 1 bit per pixel to represent either black or white. Depending on communication constraints, we may choose one mapping or switch between them based on $I(y)$ and $c(t)$ in Eq. \eqref{second map h opt}. Similarly, semantic segmentation algorithms typically output multiple channels of data, which further increases the amount of data that needs to be transmitted. Figure \ref{fig:bluesim_experiments} shows a simulated image of the underwater environment and its respective grayscale and black-and-white versions.  

In this way, the observation space of the operator $\mathcal{Y}_o$ is composed of processed video frames and can be defined as follows:

\begin{equation}
    \mathcal{Y}_r = \mathbb{R}^{M' \times N'}\times [0,T]
    % \mathcal{Y}_{r}=[0,255]\times[0,255]\times[0,255]\times [0,M] \times [0,h] \times T
\end{equation}
where $M'\leq M$ and $N'\leq N$.

One natural candidate to evaluate the performance of the proposed approaches is measuring the amount of information lost by computing those mappings. A popular concept for measuring similarity between images is mutual information, which can be thought of as the amount of information obtained from a random variable by observing another. In our case,  consider two images $X$ and $Y$; they define the probability distributions $p_X(x)$, $p_Y(y)$ and their joint distribution $p_{(X,Y)}(x,y)$. Therefore, the mutual information obtained from $Y$ by observing $X$ is defined as 

\begin{equation}
\eqsize
    \label{mutual information}
    \mathcal{I}(X,Y) = \iint_{\mathbb{R}^n \times \mathbb{R}^n} p_{(X,Y)}(x,y) \log \Bigg(\frac{p_{(X,Y)}(x,y)}{p_X(x) p_Y(y)} \Bigg)dxdy.
\end{equation}

Concretely, we could consider maximizing the functional \eqref{second map h opt} associated with the average information preserved by the transformation 

\begin{equation}
\eqsize
    K(h) =\mathbb{E}_{y\sim p(y)} [\mathcal{I}(h(y),y)  ] = \int_{\mathcal{Y}_r} p(y) \mathcal{I}(h(y),y) dy  \
\end{equation}

% $I(y)$ as the sent bit count of the image $y$ and 
$p(y)$ is a probability distribution defined on $\mathcal{Y}_r$. 

For example, the image in Fig. \ref{fig:bluesim_experiments} (top) was used to calculate mutual information between the original image and its grayscaled and black-and-white versions. The values were $2.008$ for the gray scale image against the original image, $0.16$ for the black-and-white image against the actual image, and $4.9$ for the original image against itself. In addition, other transformations for images, such as semantic segmentations~\cite{islam2020semantic} or transformations inspired by human-friendly representations for navigations~\cite{maceachren1986linear}, can be used.

\section{Implementation of the prototype}
\label{sec:implementation}

\begin{figure}[ht!]
    \centering
    \includegraphics[width=\linewidth]{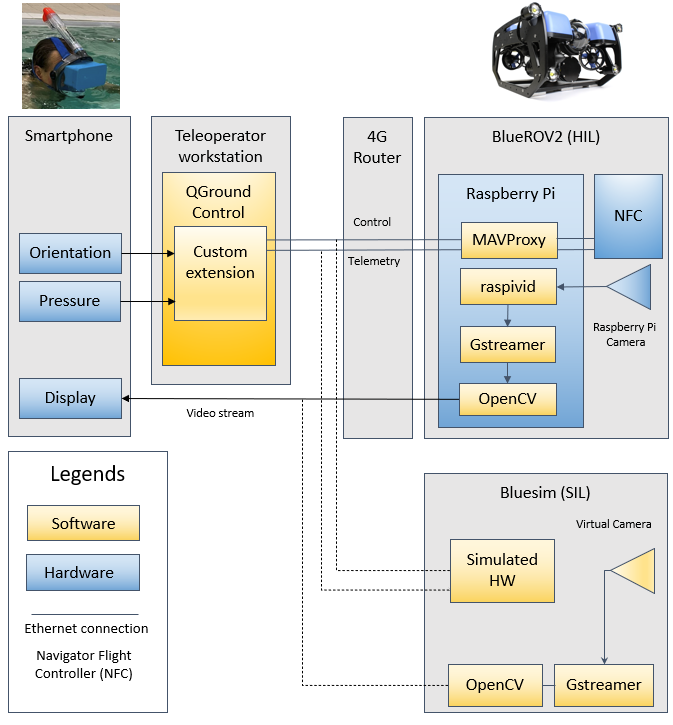}
    \caption{Architecture of the initial prototype}
    \label{fig:architecture}
\end{figure}

This work utilized two different approaches, namely Software-in-the-Loop (SIL) and Hardware-in-the-Loop (HIL). SIL configuration uses the BlueSim hardware simulator \cite{bluesim} instead of actual hardware. The SIL configuration employed the BlueSim hardware simulator \cite{bluesim} instead of real hardware for software component development and configuration. BlueSim simulates the BlueROV2 hardware \cite{bluerov2}, providing a virtual camera unit for testing and refining the system. Fig. \ref{fig:architecture} presents the architecture of the initial prototype and connections of the SIL and HIL. 

In contrast to the SIL configuration, the HIL configuration employs actual hardware to control the ROV. The ROV, located in the FIU laboratory, establishes a tethered connection with an above-surface 4G router, enabling the transmission of black-and-white video and control of the ROV's functions. To stream the underwater video, the BlueROV2 utilizes Gstreamer~\cite{Gstreamer}, and an OpenCV~\cite{OpenCV}-based application was developed to process the video stream and transmit only black-and-white images to conserve bandwidth for the teleoperator. The BlueROV2 connects to the internet via a 4G router modem, allowing for location-independent ROV teleoperation using a public static IP address. As a result, a modified version of QGroundControl (QGC) \cite{Qgroundcontrol} installed on the teleoperator's workstation allows for ROV control. Finally, a smartphone connects to the QGC, enabling the teleoperator to use body movements to control the ROV.

The teleoperating control device consists of a diving mask, casing, and smartphone. The diving mask is a standard full-face type diving mask. The smartphone casing is a custom-designed piece that transforms the diving mask and smartphone into an HMD. A point cloud of the mask was extracted using a three-dimensional laser scanner before designing the casing in 3D-CAD software to ensure an accurate fit between the mask and casing. The casing prototype was printed out of polylactide (PLA) material using a 3D printer at the Production Automation Laboratory at Centria University, Finland. 

To access the smartphone's inertial measurement unit (IMU) orientation and pressure data, we used the Sensorstream IMU+GPS application developed by Lorenz~\cite{Sensorstream}. This application reads and streams the IMU and pressure data to the UDP port of the teleoperator workstation. A custom extension was developed for the QGC to receive the sensor data stream and translate the orientation and pressure data into directional commands for the ROV, as well as up and down commands for the ROV camera.   

\section{Experimental Results}
\label{sec:results}

\subsection{Optimal Interface}

An example of a human-robot interface is shown in Fig. \ref{fig:human_robot_config}, but there could be numerous potential interface designs. We aim to show that this ``natural" interface is optimal in some sense. To this end, we concretely define the action spaces of the operator and the robot. Since the camera angle does not affect the robot's movement, we consider an operator who can perform two actions: moving their head and moving their body (e.g., by swimming or walking). These actions are denoted as $a_{head}(t)$ and $a_{body}(t)$, respectively, and we use $a(t) = [a_{head}(t), a_{body}(t)]^\top$ to represent the operator's action space, where $\mathcal{A}\subseteq \mathbb{R}^2$ and $\mathcal{C}_0 \subseteq \mathbb{R}^2$. The kinematic model of the robot is given by

\begin{equation}
\eqsize
\begin{aligned}
\dot{x}_{pos}(t) &= v(t) \cos(\theta(t))\\
\dot{y}_{pos}(t) &= v(t) \sin(\theta(t))\\
\dot{\theta}(t) &= w(t)
\end{aligned}
\end{equation}

where $v(t)$ is the forward speed and $\omega(t)$ is the angular speed. We use $u(t) = [v(t), w(t)]^\top$ and $\dot{x}(t)=[\dot{x}_{pos}(t),\dot{y}_{pos}(t), \dot{\theta}(t)]^\top$ to represent the robot's action and state spaces, respectively, where $\mathcal{U}\subseteq \mathbb{R}^2$ and $\mathcal{C}_r \subseteq \mathbb{R}^2\times S^1$.

Our objective, as described in Problem 1, is to find the map $g$. In this case, $g$ is assumed to be a linear transformation given by $u(t) = Ga(t)$ and exhibits several properties, including continuity, linearity, and consistency under certain conditions. We consider the task of moving the robot from the initial point $x_{initial}$ to the final point $x_{final}$ since this is one of the most performed tasks when environment exploration is being carried out. We aim to find the optimal interface $g$ and the control policy $a(t)$ that should be applied by the operator. To achieve this, we define the following optimization problem inspired by \eqref{arc length} and \eqref{curvature}

\begin{equation}
\eqsize
\begin{aligned}
    \label{functional 1}
    \min_{g,a} \quad & \alpha ||x_{final}-x(T)||^2 + \beta \int_0^T a(t)^\top M a(t)dt\\ 
    &+ \gamma \int_0^T ||\dot{x}(t)||dt +  \delta \: dist(G,O(2)).\\
    \text{S.t.} \quad &\dot{x}(t) = 
    \begin{bmatrix} 
    \cos(\theta(t)) & 0\\
    \sin(\theta(t)) & 0 \\ 
    0 & 1
    \end{bmatrix} u(t), \quad x(0) = x_{initial}.
\end{aligned}
\end{equation}

where $M$ is a positive-definite matrix, $O(2)$ is the set of orthogonal matrices of size $2 \times 2$, and the coefficients $\alpha, \: \beta, \: \gamma$, and $\delta$ are non-negative regularization coefficients that determine the relative importance of each term. The first term ensures that the desired point is reached, given the control policy of the operator after being transformed by the interface. %The first term ensures that the desired point is reached by the control given by the operator after being transformed by the interface. 
The second term measures the effort made by the user, with a higher cost assigned to head movements compared to body movements to keep a comfortable interface for the user. The third term considers the distance the robot traverses and encourages it to take the optimal path. The fourth term encourages linear transformation to preserve angles. Hence, $G$ should be an orthogonal matrix for it to fulfill the consistency criterion better. This term is expressed as the distance between $G$ and the set of orthogonal matrices in the Frobenius norm, which is $||U V^\top-G||_F$, where $G = U\Sigma V^\top$ is the singular value decomposition of the matrix $G$, and $||\cdot||_{F}$ is the Frobenius norm.

\newcommand{\hsp}{-0.5cm} 
\begin{figure}[th]
    \centering
    % \hspace{-0.5cm}
    \begin{tabular}{c}
    \hspace{\hsp}
        \includegraphics[width=\columnwidth]{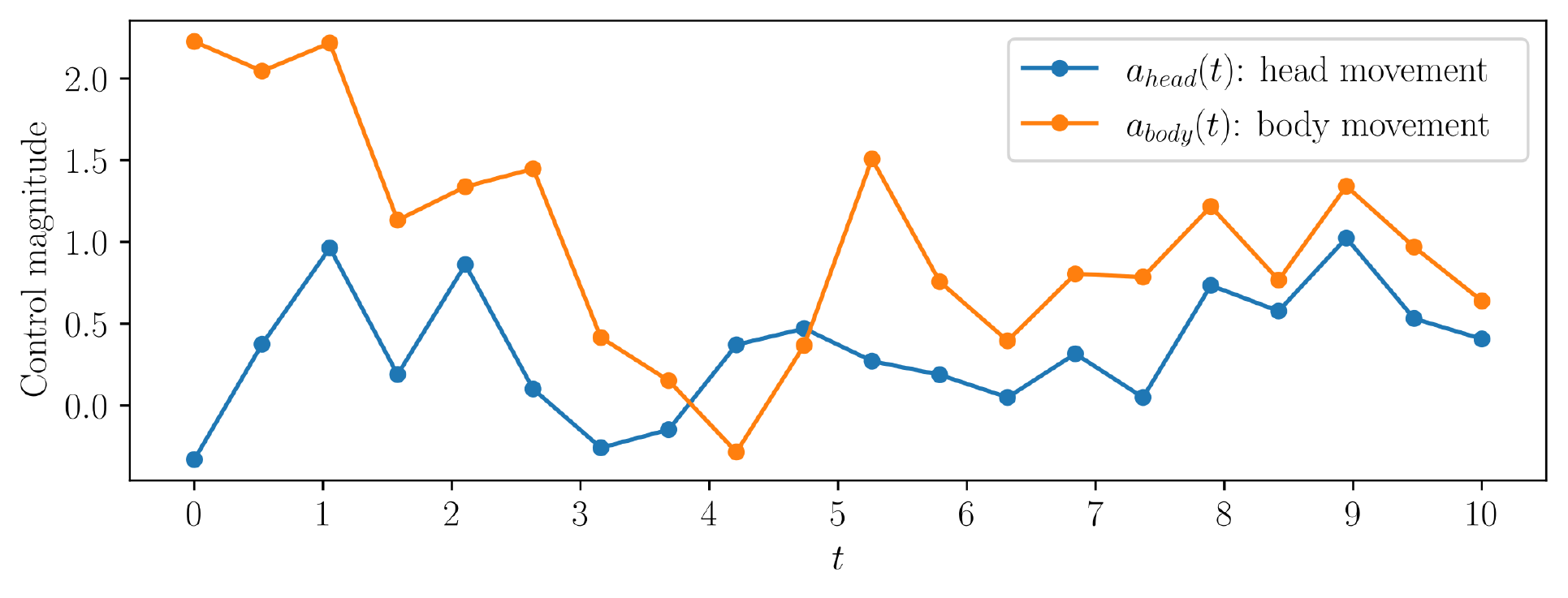}\\
        % (a)\\
        \hspace{\hsp}
       \includegraphics[width=\columnwidth]{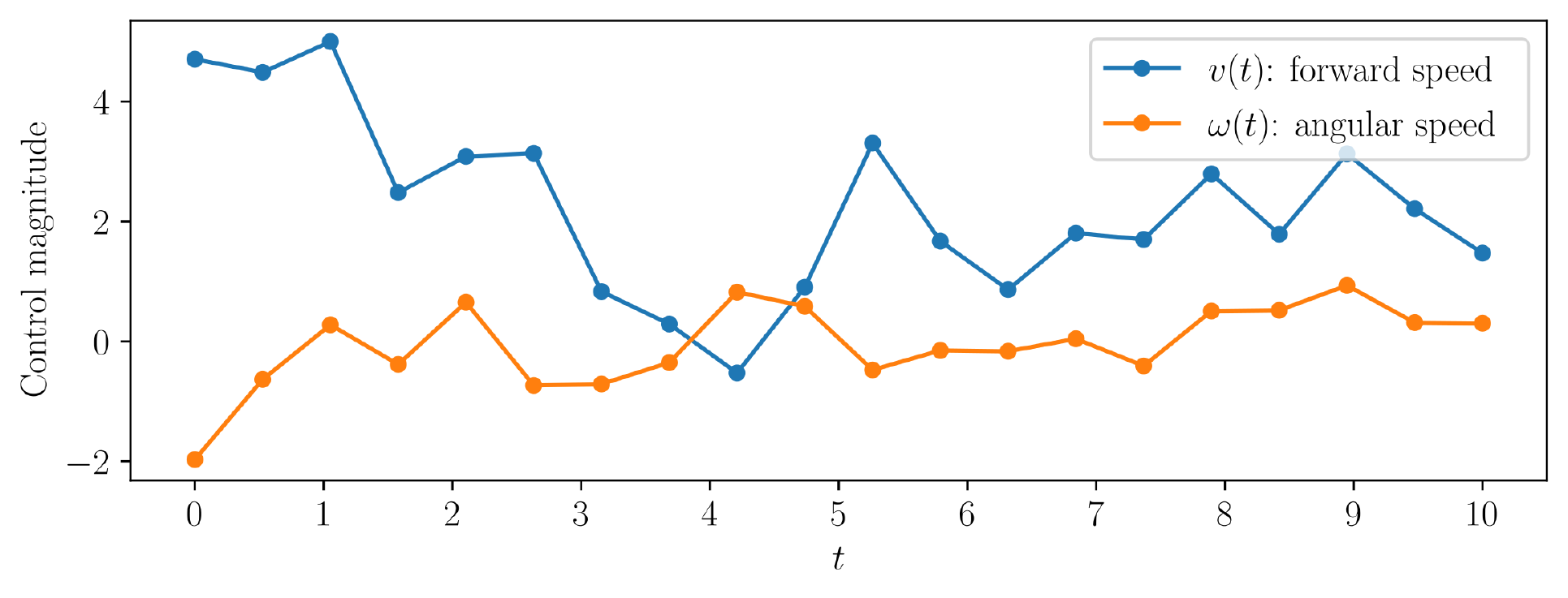}\\
        % (b)\\
        \hspace{\hsp}
        \includegraphics[width=\columnwidth]{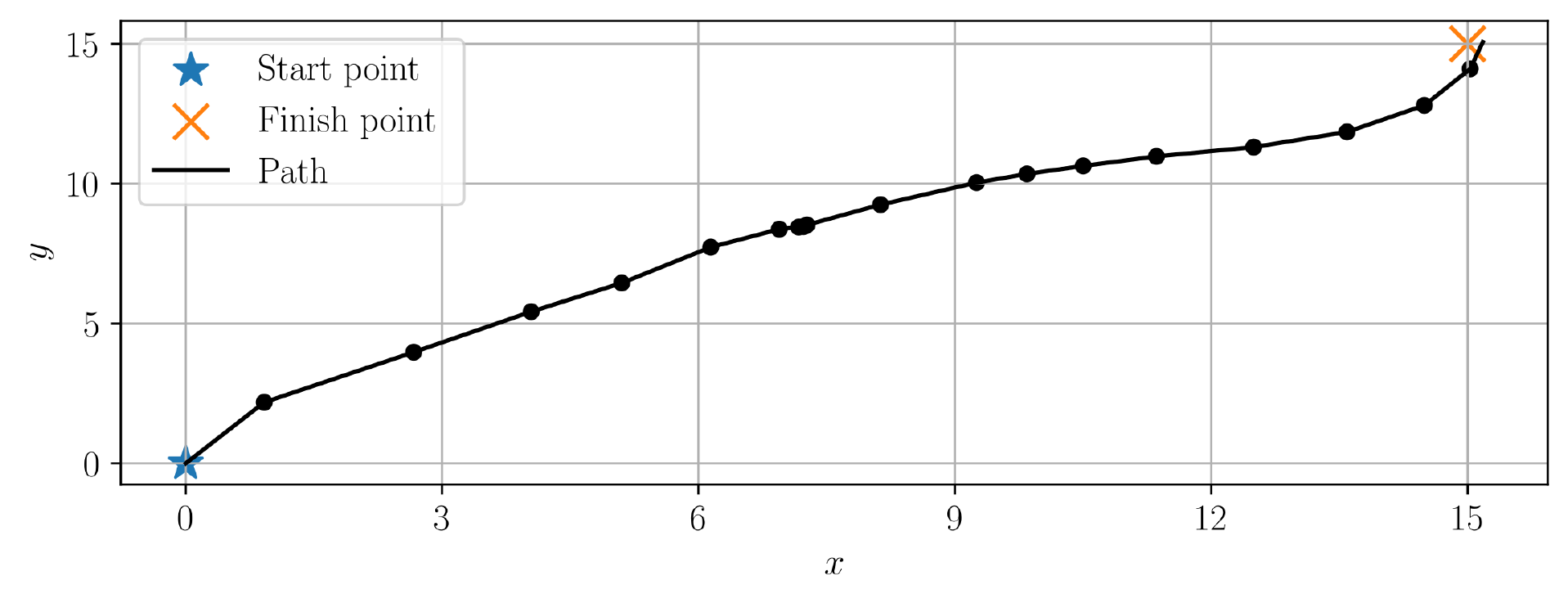}\\
        % (c)      
    \end{tabular}
    \caption{Experiment results solving problem \eqref{functional 1}. $x_{initial}= [0,0,\pi/2]^\top, \: x_{final}= [15,15,-\pi/2]^\top, \: M = \text{diag}(10,1/2), \: \alpha = 250, \: \beta =10, \gamma =5$ and $\delta = 10$. Control $a(t)$ performed by the operator (top); control $u(t)$ mapped to the robot (middle); and simulated path task (bottom).
    }
    \label{fig:optimal interface}
\end{figure}

Simulation experiments were conducted to find the optimal set of actions applied by the operator ($a(t)$, Fig \ref{fig:optimal interface} top); we noticed that instead of relying on excessive head movements, the operator can complete the assigned task by executing body movements. Also, we found the optimal interface $g$, the control applied to the robot $u(t)=Ga(t)$ ($u(t)$, Fig \ref{fig:optimal interface} middle), and the path taken by the robot ($x(t)$, Fig \ref{fig:optimal interface} bottom). More important is the map $g$, which turned out to be
%We can see that the operator does not have to use too much of his head to perform the assigned task, rather it is accomplished by performing forward movements.  

\begin{equation}
\eqsize
    \label{optimal g}
    G = \begin{bmatrix}
         0.24 & 2.15 \\ 
         1.73 & -0.62
    \end{bmatrix}
    \approx \begin{bmatrix}
         0 & 2 \\ 
         2 & 0 
    \end{bmatrix}.
\end{equation}

Assuming the coarse approximation expressed in \eqref{optimal g}, it can be observed that there exists a relationship between the human control $a(t)$ and the robot control $u(t)$ described as 
%If we allow the coarse approximation in \eqref{optimal g} for explanation purposes, we see that the human control $a(t)$ and the robot control $u(t)$ are related by 

\begin{equation}
\eqsize
\label{relation a and u}
\begin{array}{c}
    \begin{bmatrix}
        v(t)\\ \omega(t)
    \end{bmatrix}=u(t) = Ga(t) \approx
    \begin{bmatrix}
         0 & 2 \\ 
         2 & 0 
    \end{bmatrix} \begin{bmatrix}
        a_{head}(t) \\ a_{body}(t)
    \end{bmatrix}\\
    \begin{bmatrix}
        v(t)\\ \omega(t)
    \end{bmatrix} \approx 2 \begin{bmatrix}
      a_{body}(t)  \\  a_{head}(t)
    \end{bmatrix}.
\end{array}
\end{equation}

It can be inferred from \eqref{functional 1} that the "natural" mapping, in which the operator directs the robot's movement by their head orientation and the robot's forward motion by their own forward movement, is not only intuitive and user-friendly but also optimal.

\begin{figure}[ht]
    \centering
    % \hspace{-0.5cm}
    \begin{tabular}{c}
    \hspace{\hsp}
        \includegraphics[scale=0.28]{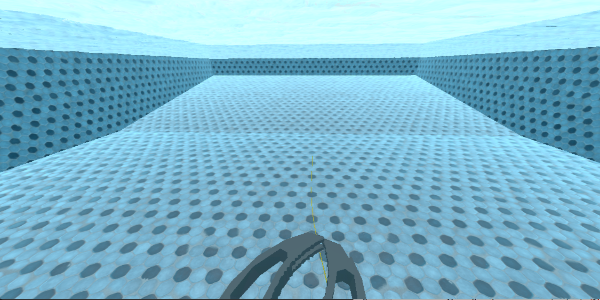}\\
        % (a)\\
        \hspace{\hsp}
        \includegraphics[scale=0.28]{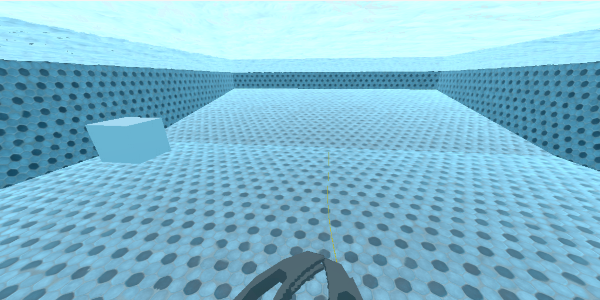}\\
        % (b)\\
        \hspace{\hsp}
        \includegraphics[scale=0.28]{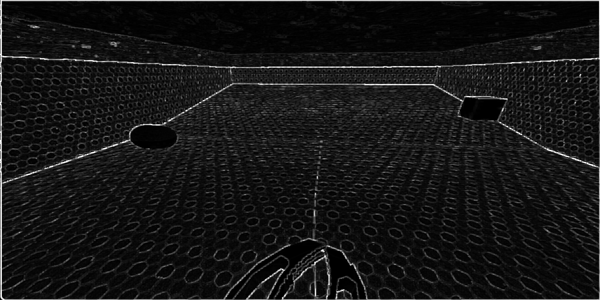}\\
        % (c)      
    \end{tabular}
    \caption{Experimental procedure for underwater teleoperation simulation. Task 1: familiarization with empty pool (top); Task 2: find cubic shape provided RGB video stream (middle); Task 3: find oval shape provided edge-detected, black-and-white video stream (bottom).}
    \label{fig:bluesim_experiments}
\end{figure}

To gather quantifiable data on the effectiveness of the presented solution, the framework is tested by a group of volunteers based on the guidelines of \cite{lavalle2019}. The experimental procedure is described in three tasks. In the first task, each user is provided an empty pool scenario and has 3 minutes for familiarization with the headset and simulator (Fig. \ref{fig:bluesim_experiments}, top). In the second task, the user is provided an RGB video stream from the robot front camera and is asked to identify a cubic shape in the pool by pointing the robot camera to the respective shape. Only one cubic shape is in one corner of the pool (Fig. \ref{fig:bluesim_experiments}, middle). In the third task, the user is provided a black-and-white video stream of the pool and is asked to identify an oval shape by pointing the robot camera to the respective shape (Fig. \ref{fig:bluesim_experiments}, bottom). In this step, there is an oval shape in one of the pool's corners and a cubic shape in another. Regarding implementing the user commands strategy, we used a fixed-size buffer to collect enough samples at the beginning of the simulation and avoid unnecessary rotations of the simulated ROV. To move the ROV, we compute the moving average of the buffer, where each sample represents the difference between two consecutive measurements of head orientations. The resulting moving average of each timestep is then checked against a predefined threshold to decide which way (left or right) the ROV should turn. The time required to complete tasks 2 and 3 is recorded during the piloting. 

\begin{figure}[ht!]
    \centering
    \includegraphics[scale=0.28]{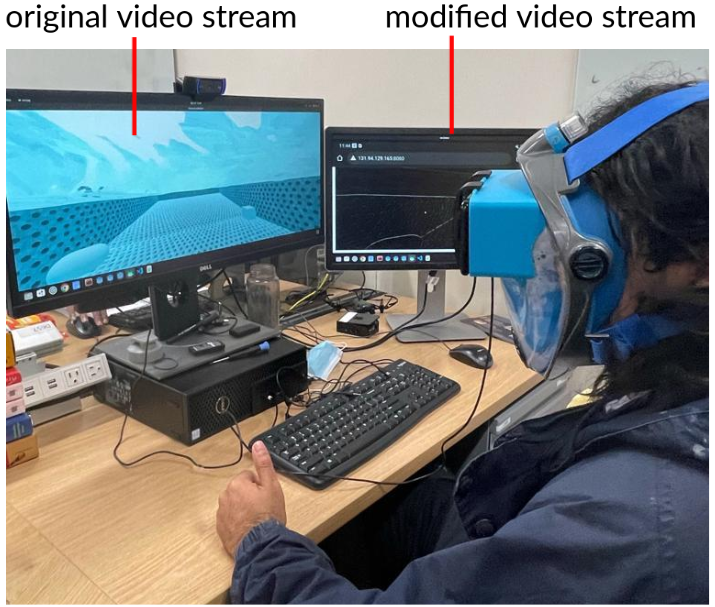}
    \caption{Experimental setup. The original video stream is displayed to the user's HMD for tasks 1 and 2, and the modified video stream is provided for task 3.}
    \label{fig:user_experiment}
\end{figure}

The group of users exhibited a balanced distribution with respect to gender, comprising three females and three males, with a mean age of 24.5 years (Fig. \ref{fig:user_experiment}). Regarding the head-mounted display, the experiments indicate that optimal teleoperation comfort can be achieved by increasing the distance between the eyes and the smartphone. Additionally, we noted that laggy communication significantly increases task completion times, as users must wait for image updates on their phone screens. As expected, users could detect shapes faster when presented with RGB streams, while black-and-white streams resulted in comparatively longer detection times. The average completion time and standard deviation for tasks 2 and 3 were $\approx 14.77 \pm 5.18$  seconds and $\approx 32.44 \pm 5.62$ seconds, respectively.

\section{Conclusions and Future Work}
\label{sec:conclusions}
%The presented approach to teleoperating an ROV with teleoperator body and head movements is a proof-of-concept. 
In this paper, we investigated the design of human-robot interfaces for robotics teleoperation based on key concepts such as linearity, consistency, continuity, and user comfort. As a proof-of-concept, the proposed solution is applied to perform underwater inspection tasks from a safe and ergonomic location (e.g. a lab environment). 
A prototype was built to enable location-independent teleoperation of an ROV utilizing the human body as a controlling device. Experimental results were performed on a group of volunteers to gather quantifiable data on the effectiveness of the presented solution. Two weaknesses in the presented architecture require redesigning. Firstly, the absence of cybersecurity strategies such as user authentication and encryption of control and video data renders the architecture vulnerable to data capture and unauthorized control of the ROV. Secondly, the control data communication method is unsuitable for long-distance communication with latencies and disconnections. Nonetheless, the prototype serves as a basis for realizing the need for a \textit{cybersecure application that uses open communication standards to facilitate safe teleoperation of an ROV}. Another direction for future work is to expand the proposed action space and configuration space mappings to account for the vehicle's depth control, translational and rotational motion, and robotic gripper operation. Finally, a third avenue for future steps is to continue our previous work on digital twins \cite{kaarlela2022} and utilize extended reality as an educational training tool for underwater robotics.

\section*{Acknowledgements} 
This research was funded by the European Research Council (ERC) under the European Union's Horizon 2020 research and innovation program (grant agreement n\textdegree{} 825196), the NSF grants IIS-2034123, IIS-2024733, and by the U.S. Dept. of Homeland Security 2017-ST-062000002.

\bibliographystyle{abbrv}
\bibliography{main}
\end{document}